\documentclass[a4paper,twoside]{article}

\usepackage{epsfig}
\usepackage{subcaption}
\usepackage{calc}
\usepackage{amssymb}
\usepackage{amstext}
\usepackage{amsmath}
\usepackage{amsthm}
\usepackage{multicol}
\usepackage{pslatex}
\usepackage{apalike}
\usepackage{algorithm2e}
\usepackage[bottom]{footmisc}

\usepackage{SCITEPRESS}     % Please add other packages that you may need BEFORE the SCITEPRESS.sty package.

\begin{document}

\title{Combi-CAM: A Novel Multi-Layer Approach for Explainable Image Geolocalization}

\author{\authorname{D. Faget\sup{1}\orcidAuthor{0009-0007-2520-3575}, J.~L.~Lisani\sup{2}\orcidAuthor{0000-0002-7004-2252} and M. Colom\sup{1}\orcidAuthor{0000-0003-2636-0656}}
\affiliation{\sup{1}Centre Borelli, ENS Paris-Saclay, Université de Paris, CNRS, INSERM, SSA, France}
\affiliation{\sup{2}Universitat de les Illes Balears, IAC3, Spain}
\email{joseluis.lisani@uib.es, miguel.colom-barco@ens-paris-saclay.fr}
}


\abstract{Planet-scale photo geolocalization involves the intricate task of estimating the geographic location depicted in an image purely based on its visual features. While deep learning models, particularly convolutional neural networks (CNNs), have significantly advanced this field, understanding the reasoning behind their predictions remains challenging. In this paper, we present Combi-CAM, a novel method that enhances the explainability of CNN-based geolocalization models by combining gradient-weighted class activation maps obtained from several layers of the network architecture, rather than using only information from the deepest layer as is typically done. This approach provides a more detailed understanding of how different image features contribute to the model's decisions, offering deeper insights than the traditional approaches.}

\onecolumn \maketitle \normalsize \setcounter{footnote}{0} \vfill

\section{\uppercase{Introduction}}
\label{intro}

\begin{figure}[!htb]
\includegraphics[width=0.32\linewidth, height=0.4\linewidth]{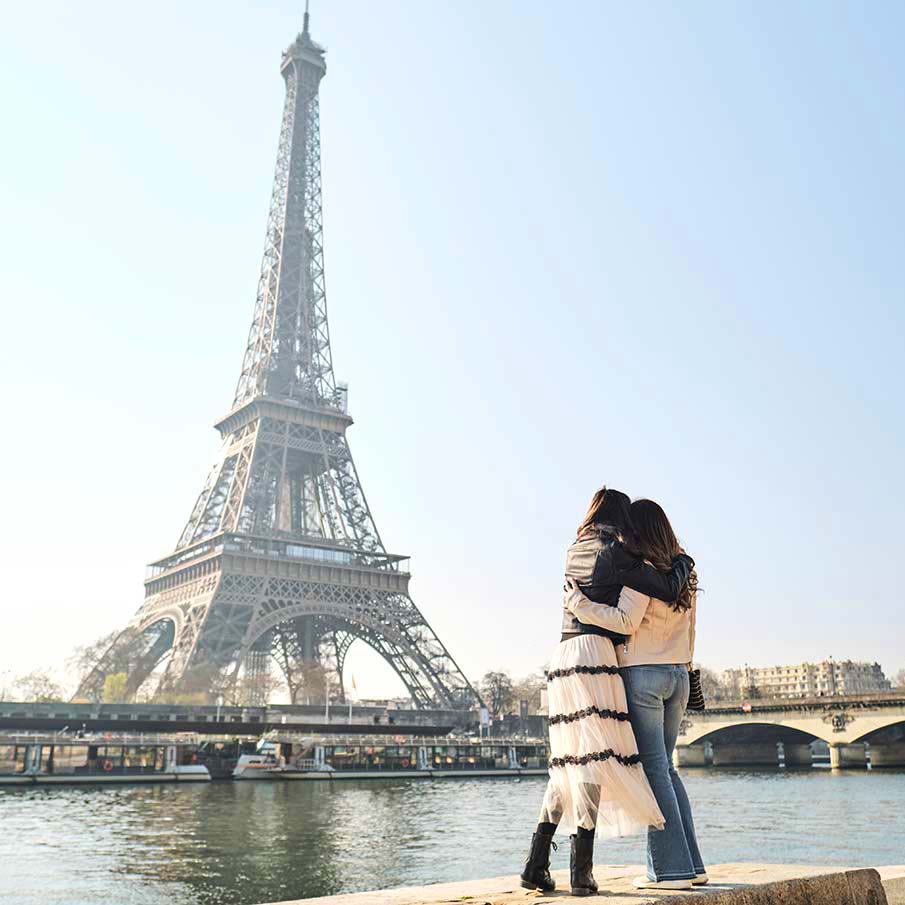}
\includegraphics[width=0.32\linewidth, height=0.4\linewidth]{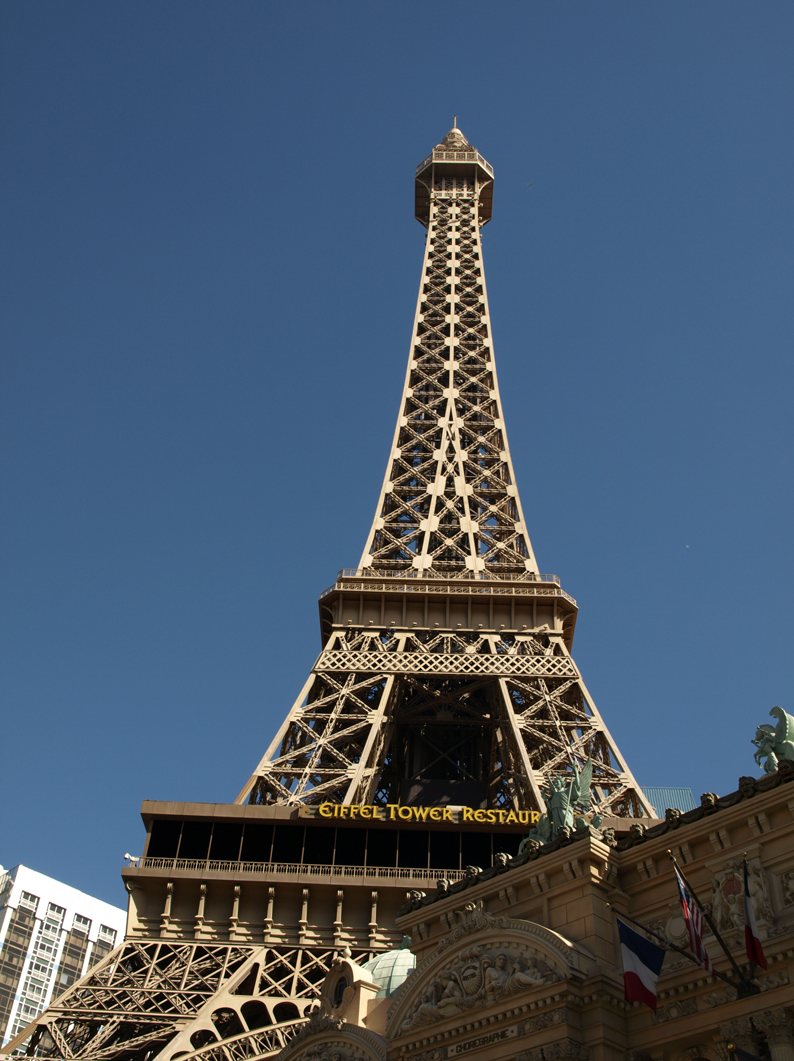}
\includegraphics[width=0.32\linewidth, height=0.4\linewidth]{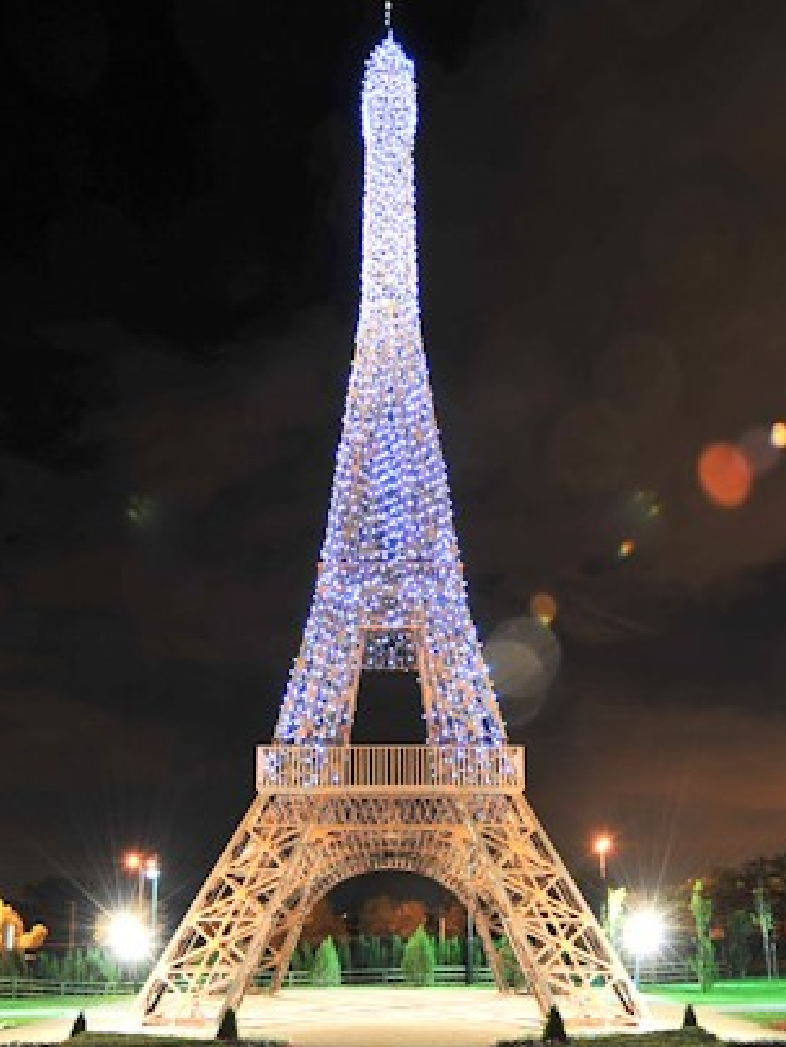}
\caption{The Eiffel Tower and its replicas: Original in Paris (left), and replicas in Las Vegas (center) and Madrid (right).}
\label{fig:intro}
\end{figure}

Geolocalization from a single image is a challenging and complex problem. Consider the three images presented in Figure~\ref{fig:intro}. At first glance, it seems obvious that all three were taken in Paris, France. However, upon closer inspection, significant differences become apparent. In reality, only the image on the left shows the original Eiffel Tower, while the one on the center is situated in Las Vegas, and the one on the right is in Madrid.
%While they may appear similar at first glance, distinct differences emerge when considering the surrounding vegetation and objects, highlighting the difficulty of the geolocalization task.

Unlike many other vision applications, single-image geolocalization often relies on fine-grained visual cues found in small regions of an image~\cite{im2gps}, ~\cite{planet}, ~\cite{cplanet}. Although the three images in Figure~\ref{fig:intro} appear to depict the same location, the buildings and vegetation in the background are crucial for distinguishing them. Similarly, for most other images, the global context covering the entire image is as significant for geolocalization as individual foreground objects. Moreover, the same location exhibits drastic appearance variations under different daytime or weather conditions.

Because of this complexity, the need for robust explainability tools becomes increasingly important. These tools provide insights into the underlying decision-making process of geolocalization models. By emphasizing fine-grained visual cues (such as the architectural style, vegetation, and any background elements) explainability tools help understand how these subtle differences contribute to model predictions. This is especially critical in scenarios where similar landmarks exist across different locations, as it helps distinguish between visually alike yet geographically distinct places. Moreover, explainability enhances the reliability of geolocalization models by ensuring that their decisions are based on meaningful and contextually relevant features, thereby enabling more transparent and trustworthy applications in real-world settings. 

In recent years, Grad-CAM \cite{gradcam} (Gradient-weighted Class Activation Mapping) has emerged as the most widely used gradient-based explainability tool in the field of computer vision and its application has been extensive in geoscience and remote sensing research \cite{remote1}, ~\cite{remote2}, ~\cite{remote3}. Grad-CAM works by highlighting the regions in an image that are most relevant to the model's predictions, typically by visualizing the gradients at the last convolutional layer of a Convolutional Neural Network (CNN). Researchers often apply Grad-CAM to this final layer as it captures the most abstract and high-level features of the input image. However, it is well known that in a CNN, each layer plays a relevant role in processing and transforming the input data \cite{cnn1}, ~\cite{cnn2}, ~\cite{cnn3}. Earlier layers capture low-level features such as edges and textures, while deeper layers progressively combine these features to encode more complex patterns and shapes.  Previous studies~\cite{interpre} have observed that, in some cases, Grad-CAM applied to the last layer may not precisely highlight key regions of an image. The authors suggest that this ``may be caused by the issue of gradient discontinuity'', although this explanation remains open to interpretation.

In this paper, we propose a novel approach that leverages activation maps throughout several layers of the network, integrating the information from each layer into a unified heatmap. By doing so, we capture not only the high-level abstractions but also the fine-grained details and intermediate patterns that contribute to the model's decision-making process. This comprehensive approach provides a more complete and explainable visualization of how different regions of an image influence the model's predictions, allowing for a deeper understanding of the network's internal working.
We show that Grad-CAM consistently highlights image regions containing key elements and occasionally also does so in the intermediate layers of the network and not only the final one. We analyze the magnitudes of pixel activations across the layers to demonstrate that, in several cases, middle layers are more informative than the final one. We apply this new technique to the geolocalization task because it is highly visual and involves integrating elements from multiple parts of the image, but it has potential for use in other tasks such as scene understanding~\cite{scene}, image segmentation~\cite{segment}, and diverse fields including medical imaging~\cite{medical} and autonomous driving~\cite{drive}, where accurate interpretation of computer vision models is essential.

The paper is organized as follows. We explain Grad-CAM and some of its variants in Section~\ref{gradcam} and describe the existing explainability methods for geolocalization in Section~\ref{inter}. We present our proposed approach in Section~\ref{new}, and Section~\ref{experiments} reports visual results. Finally, Section~\ref{conclusion} concludes the article.

\section{\uppercase{Grad-CAM: an interpretability tool}}
\label{gradcam}

Grad-CAM~\cite{gradcam} is a popular explainability method used to visualize the contribution of different regions in an image to the final decision of a convolutional neural network. This method highlights the relevant areas in the input image that most influence the network's prediction.

To understand Grad-CAM, let \( A^\lambda_k \) represent the activation of the \( k \)-th feature map from the convolutional layer $\lambda$. For a given class \( c \), the goal is to determine the importance of each feature map \( A^\lambda_k \) in contributing to the class score \( y^c \).

The importance weights \( \alpha^c_k \), which reflect the contribution of the \( k \)-th feature map to $y^c$, are computed by averaging the gradients of $y^c$ with respect to the activations \( A^\lambda_k \):

\begin{equation}
\alpha^{\{\lambda, c\}}_k = \frac{1}{Z} \sum_{i} \sum_{j} \frac{\partial y^c}{\partial A^\lambda_{k_{i,j}}},
\label{eq:gradients}
\end{equation}

where \( Z \) is the number of pixels in each feature map of layer $\lambda$, and \( \frac{\partial y^c}{\partial A^\lambda_{k_{i,j}}} \) denotes the partial derivative of the class score with respect to the activation at spatial location \( (i, j) \) in the feature map \( A^\lambda_k \).

Finally, the class-specific localization map for the layer \( L^{\{\lambda, c\}} \) (or heatmap) is computed by applying a ReLU activation to the weighted sum of the feature maps, using the importance weights \( \alpha^{\{\lambda, c\}}_k \):

\begin{equation}
L^{\{\lambda, c\}} = \text{ReLU}\left(\sum_{k} \alpha^{\{\lambda, c\}}_k \cdot A^\lambda_k\right).
\label{eq:heatmap}
\end{equation}

This heatmap \( L^{\{\lambda, c\}} \) highlights the regions in the input image that most strongly influence the prediction of class \( c \), by emphasizing the positive contributions of each of the feature maps.
Although Grad-CAM can be computed for any convolutional layer, it is typically applied only on the last layer of the network.

Several variants of Grad-CAM have been proposed. Grad-CAM++ \cite{gradcam++} improves upon the basic Grad-CAM by addressing its limitations in localization and interpretation. It incorporates higher-order derivatives to better handle cases where multiple regions contribute to the class score, leading to more precise and detailed activation maps. Score-CAM~\cite{scorecam} proposes a more accurate method to assess the contribution of each location in the feature map which is not based on gradient information. 
Similarly to Grad-CAM, these two variants are applied exclusively to the final convolutional layer of the network, but recent works take into account the activations of shallower layers. Layer-CAM~\cite{layercam} computes the contribution of each pixel in the prediction as the maximum of the activation maps obtained at different stages of the network architecture. CSG-CAM~\cite{CSG-CAM} weights the activation maps using channel saliency and gradients and combines information from shallow and last convolutional layers. 

Our proposal is based on ideas similar to~\cite{layercam}, but the integration of the information of the different network's stages is done differently and the obtained results are more accurate in terms of localization of regions of interest in the images.

\section{\uppercase{Interpretability in Geolocalization Networks}}
\label{inter}

Given the complexity of the geolocalization task, a robust explainability tool is essential both for understanding the reasoning behind predictions and for improving network performance. Existing geo-AI research~\cite{translocator}, ~\cite{interpre} utilize Grad-CAM applied to the network's last layer for interpretability. However, Hsu observed that Grad-CAM sometimes struggles to accurately detect object shapes~\cite{interpre}. Our findings support this observation, but only for the last layer. While important objects are consistently detected in earlier layers, the last layer often focuses on a broader context. When the task requires integrating information from various points in the image, which is the case for geolocalization, applying Grad-CAM to the last layer alone may fail to accurately capture object shapes.

We focus on the geolocalization network developed by Kordopatis-Zilos~\cite{certh} and investigate its behavior by applying Grad-CAM to the last layer of every block of its EfficientNet-b4 \cite{efficientnet} backbone (which has 32 blocks, or stages, composed of several layers). 
%rather than just the last layer of the last block. 
Our primary observation is that the results obtained from applying Grad-CAM to the last layer of the last block are never the most intuitive or informative. On the other hand, applying this tool to the last layer of earlier blocks consistently provides better localization of important objects. This suggests that the feature vector obtained after processing the image in the CNN is more influenced by intermediate layers than by the final one. This is especially true if objects whose size is within the receptive of view of the neurons in that layer triggered a strong detection response. Normally this response is unique to a particular stage and layer of the network. While it is true that the final layer of the final stage of the CNN is used to produce the final decision, the particular weights it contains where generated from previous activations due to detections. This is the reason behind considering not only the last, but all the stages in the network's processing pipeline.

%This can be attributed to how information is processed in the CNN's pipeline: initial layers focus on detecting textures, subsequent layers identify objects, and the final layers integrate broader contextual information. Therefore, the activation maps generated from the last layers tend to highlight this broader context rather than specific objects. However, for explainability purposes, it is often more informative to focus on the layers where objects are clearly identified, as this provides better insights into the model's decision-making process.

%It is also important to note that features not highlighted in the last layer remain relevant.
When analyzing the output from the last convolutional layer in a particular stage of the pipeline, we access a comprehensive representation of all features processed across the convolutional layers. This includes features that may not be clearly visible in the decision-making layers as highlighted by Grad-CAM. Even if these features are not dominant in the final stages of the network, they significantly influence the overall feature vector, affecting both classification and retrieval tasks. This demonstrates how neural networks utilize a broader array of features, beyond those most visible in the final decision-making stages, for tasks such as geolocalization.

\begin{figure}[!htb]
    \centering
\includegraphics[width=0.8\linewidth]{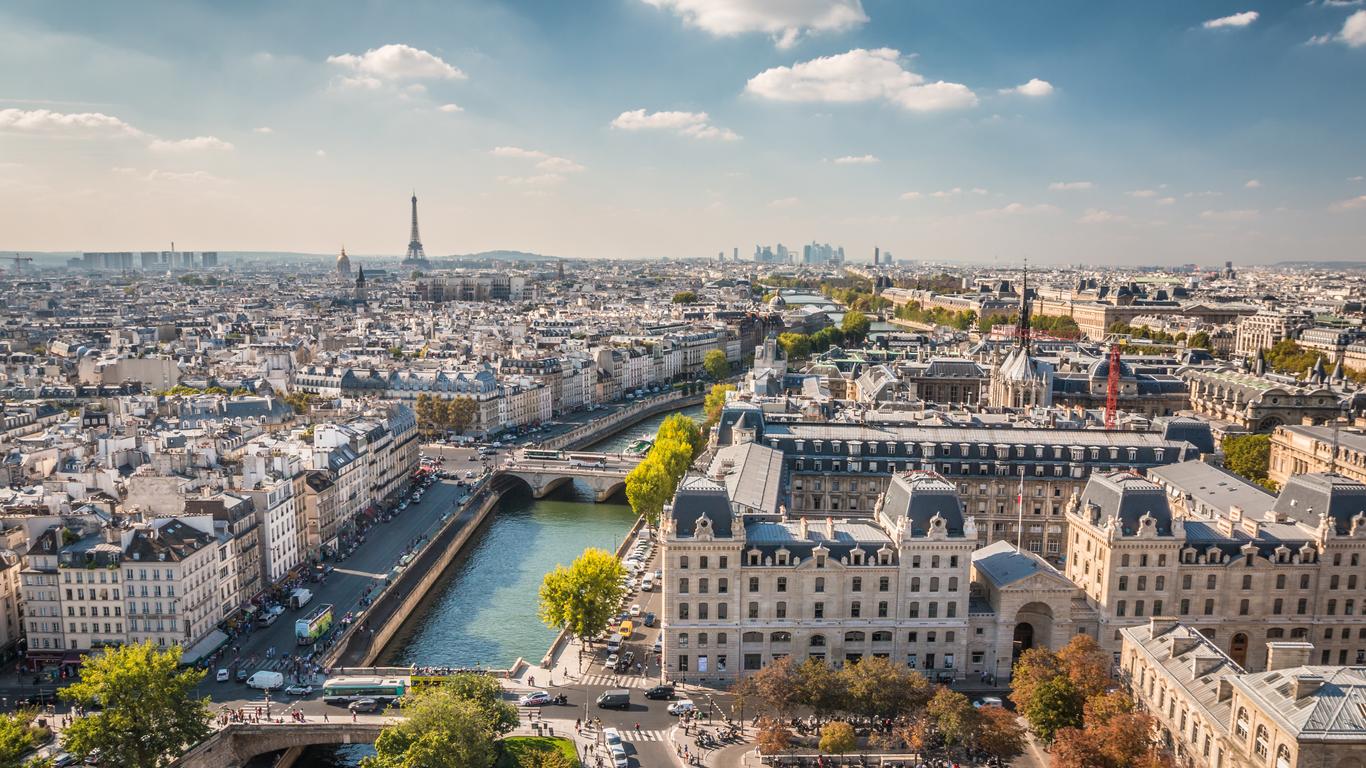}
\caption{Aerial image of Paris (France) showing a wide outlook of the city. It includes characteristic objects such as the Eiffel Tower, the Seine river, and buildings with particular architectural elements.}
\label{fig:paris}
\end{figure}

\begin{figure}[!htb]
    \centering
    % first row
    \subfloat[{\footnotesize Block \#7}]{\includegraphics[width=0.47\linewidth]{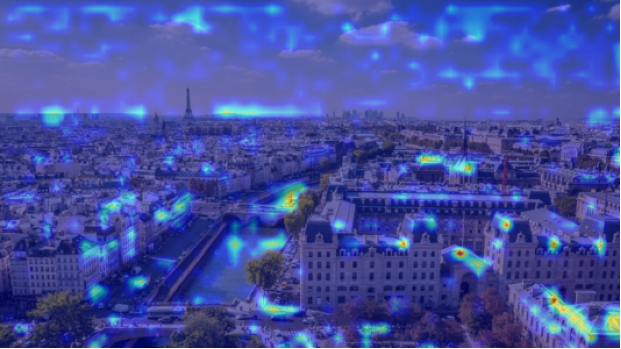}}\hspace{0.02\linewidth}%
    \subfloat[{\footnotesize Block \#15}]{\includegraphics[width=0.47\linewidth]{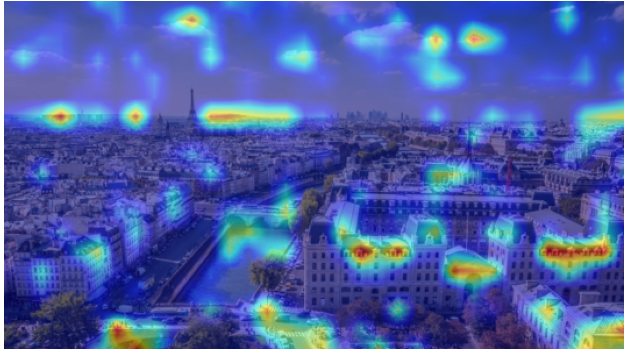}}\\[0.2cm] % increase vertical space
    % second row
    \subfloat[{\footnotesize Block \#22}]{\includegraphics[width=0.47\linewidth]{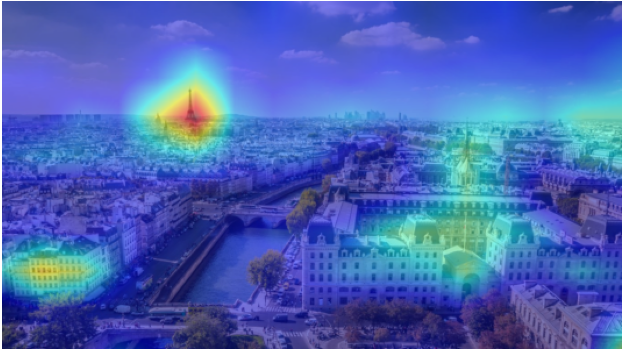}}\hspace{0.02\linewidth}%
    \subfloat[{\footnotesize Block \#28}]{\includegraphics[width=0.47\linewidth]{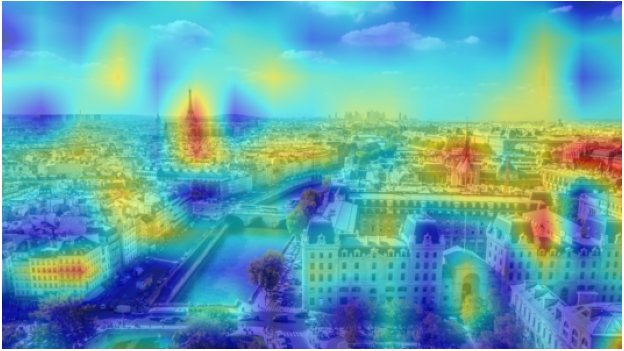}}\\[0.2cm]
    % third row
    \subfloat[{\footnotesize Block \#30}]{\includegraphics[width=0.47\linewidth]{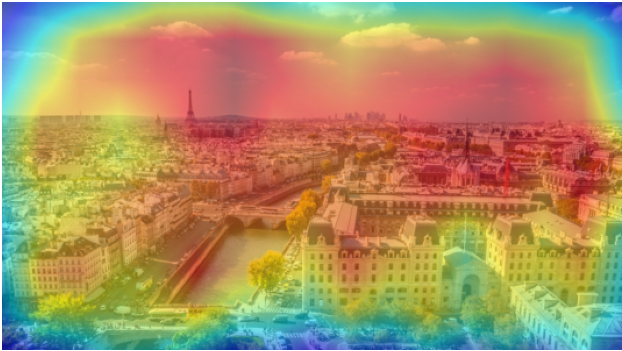}}\hspace{0.02\linewidth}%
    \subfloat[{\footnotesize Block \#31 (classic Grad-CAM output)}]{\includegraphics[width=0.47\linewidth]{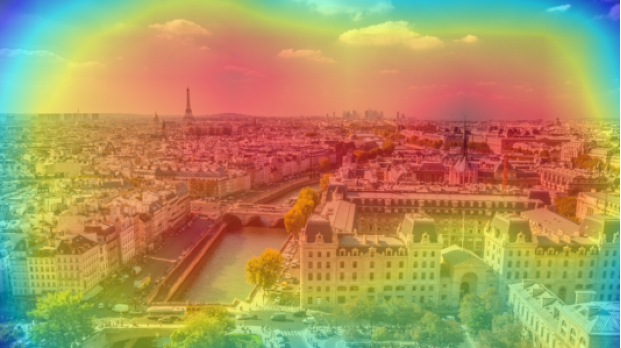}}\\[0.2cm]
    
    \caption{Results obtained by applying Grad-CAM on the last layer of selected blocks (from \#0 to \#31) of the EfficientNet-B4 architecture show that characteristic elements, such as the Eiffel Tower, become more prominent in the middle blocks of the analysis compared to the final blocks.}
    \label{fig:heatmaps}
\end{figure}

In Figure~\ref{fig:heatmaps} we present an analysis of an aerial image of Paris (Figure~\ref{fig:paris}) using Grad-CAM applied to the final layer of various blocks of the EfficientNet-B4 model. The activation maps obtained are displayed in the figure in the form of heatmaps, where the highest values are represented in red and the lowest values in blue. One could expect that the geolocalization network will identify the Eiffel Tower, a prominent landmark of Paris. However, as observed, the Eiffel Tower is highlighted only from block 22 onward. Additionally, we note that the area covered by the maps increases with deeper layers. This expansion is attributed to the architecture of the EfficientNet-B4 model, particularly its backbone and downsampling layers. A larger receptive field in the final layers is important for effective contrastive learning~\cite{certh}.

\begin{figure}[!htb]
    \centering
\includegraphics[width=0.9\linewidth]{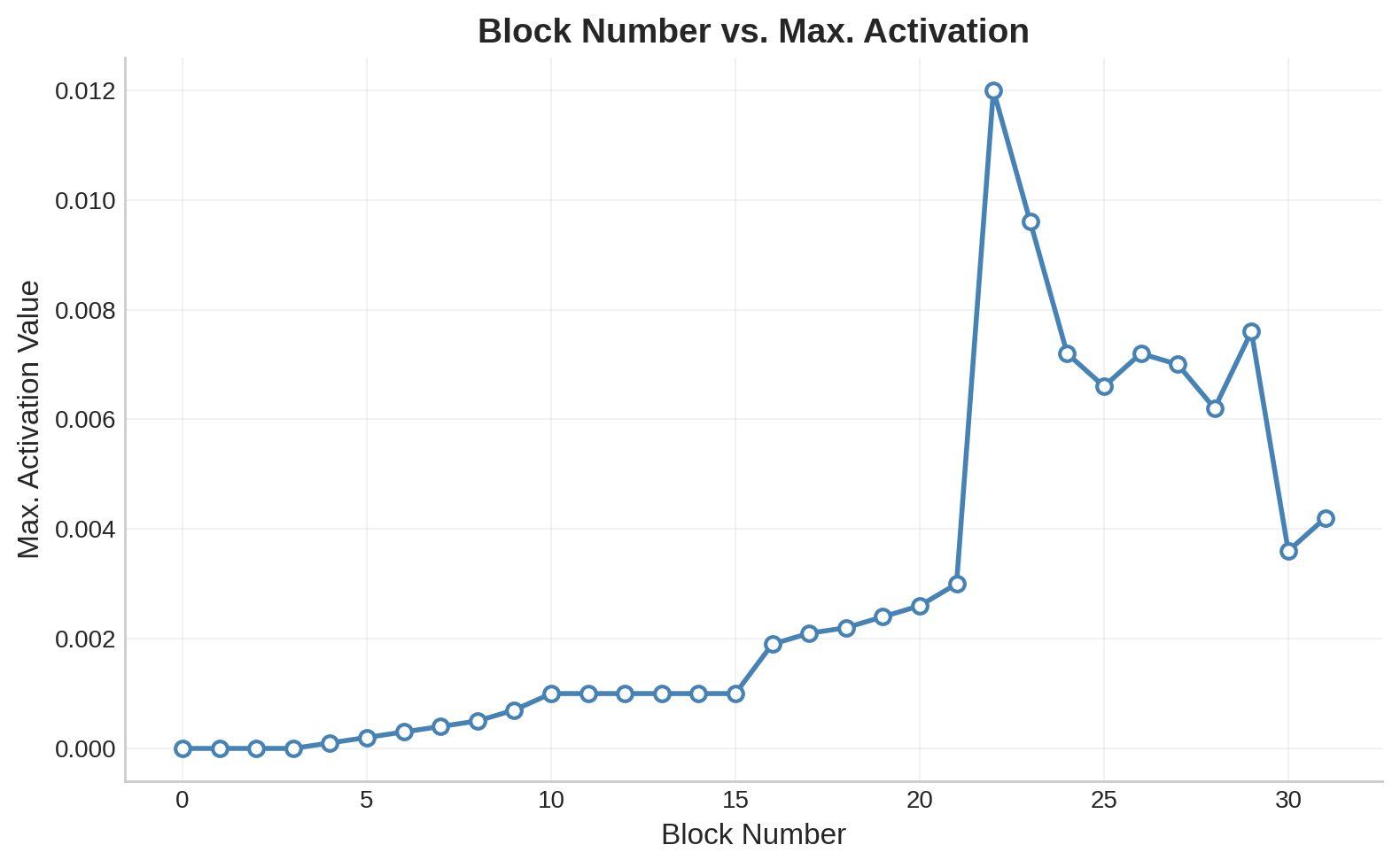}
\caption{The maximum activation magnitude per block indicates that the highest activations occur in blocks \#22 to \#29, where the pixels of characteristic elements activate the most. More specifically, the most significant block is the 22 which prominently highlights the most recognizable element of Paris: the Eiffel Tower.}
\label{fig:activations}
\end{figure}

It should be noted that the heatmaps shown in Figure~\ref{fig:heatmaps} are normalized, which means that the maximum value is always displayed in red and the minimum in blue, independently of their magnitude. Therefore, they reflect which areas are most activated in the final layer of each block but they do not allow for a valid comparison between blocks. Figure~\ref{fig:activations} illustrates this fact. It displays the maximum activation magnitudes of the last layer of each block of the EfficientNet-B4 architecture for the input image~\ref{fig:paris}. Since the maximum at each layer is different, we must be careful when comparing the heatmaps in Figure~\ref{fig:heatmaps}. For instance, the red color for the last layer of block \#30 represents a magnitude much smaller than the red color for last layer of block \#22. As anticipated, the peak activation is observed when Grad-CAM concentrates on the Eiffel Tower in block \#22. However, the activation in the final layer is notably lower, indicating that blocks \#22 through \#29 contribute more significantly to the final decision and the formation of the feature vector than the last block does.

\section{\uppercase{Combi-CAM}}
\label{new}

Let \(f\) be a CNN, let \(y^c(\mathbf{x})\) denote the class score for class \(c\) given input image \(\mathbf{x} \in \mathbb{R}^{H \times W \times 3} \), and let \(A_{\lambda}^k \in \mathbb{R}^{H_{\lambda} \times W_{\lambda}}\) be the \(k\)-th feature map at a convolutional layer \(\lambda\). In this part, we use the same symbols and conventions as in Section \ref{gradcam}.

Because spatial resolutions may differ across stages, it is convenient to introduce an upsampling operator \(\mathcal{U}_{\lambda}:\mathbb{R}^{H_{\lambda}\times W_{\lambda}}\to\mathbb{R}^{H\times W}\) (e.g., bilinear interpolation) to align all maps to the input image spatial dimensions before aggregation. Let \(\Lambda\) denote the set of last convolutional layers of the blocks of the network, and define the multi-layer Combi-CAM heatmap by
\begin{equation}
L^{c} \;=\; \sum_{\lambda \in \Lambda} \mathcal{U}_{\lambda}\!\big(L^{\{\lambda, c\}}\big),
\label{eq:combicam}
\end{equation}
where $L^{\{\lambda, c\}}$ is defined as in Equation~\eqref{eq:heatmap}, with \(\operatorname{ReLU}\) restricting the aggregation to features that have a positive influence on \(y^c\). This formulation yields a class-specific explanation that integrates fine-grained and high-level evidence.

When \(\Lambda\) contains a single layer—typically the last convolutional layer—Combi-CAM reduces to the standard Grad-CAM map $L^{\{\lambda, c\}}$, showing that the proposal strictly generalizes Grad-CAM while retaining its interpretive semantics. Aggregating across layers preserves nonnegativity due to the ReLU and enriches the explanation by accumulating complementary evidence captured at distinct representational depths.

In this work, \(\Lambda\) is instantiated as the set of last convolutional layers from all 32 blocks of the EfficientNet-B4 backbone used for geolocalization in ~\cite{certh}. However, the above procedure can be applied in general to any architecture composed of several stages (e.g. VGG16).
Qualitative analyses presented in next section indicate that intermediate blocks often deliver sharper, more localized evidence, whereas deeper blocks contribute broader semantic context, and Combi-CAM combines these advantages in a single heatmap.

\section{\uppercase{Experiments and Discussion}}
\label{experiments}

Given an input picture, the geolocalization network~\cite{certh} generates the longitude and latitude coordinates of the most likely location where it could have been taken. For example, for the image shown in Figure~\ref{fig:paris} the location is Paris (France), with a confidence score of 100\%. We applied state-of-the-art explainability methods (Grad-CAM, Grad-CAM++, Layer-CAM and Combi-CAM) to understand the output of the network and displayed the results in Figure~\ref{fig:CAMs}-left. 

\begin{figure}[!htb]
    \centering
\includegraphics[width=0.7\linewidth]{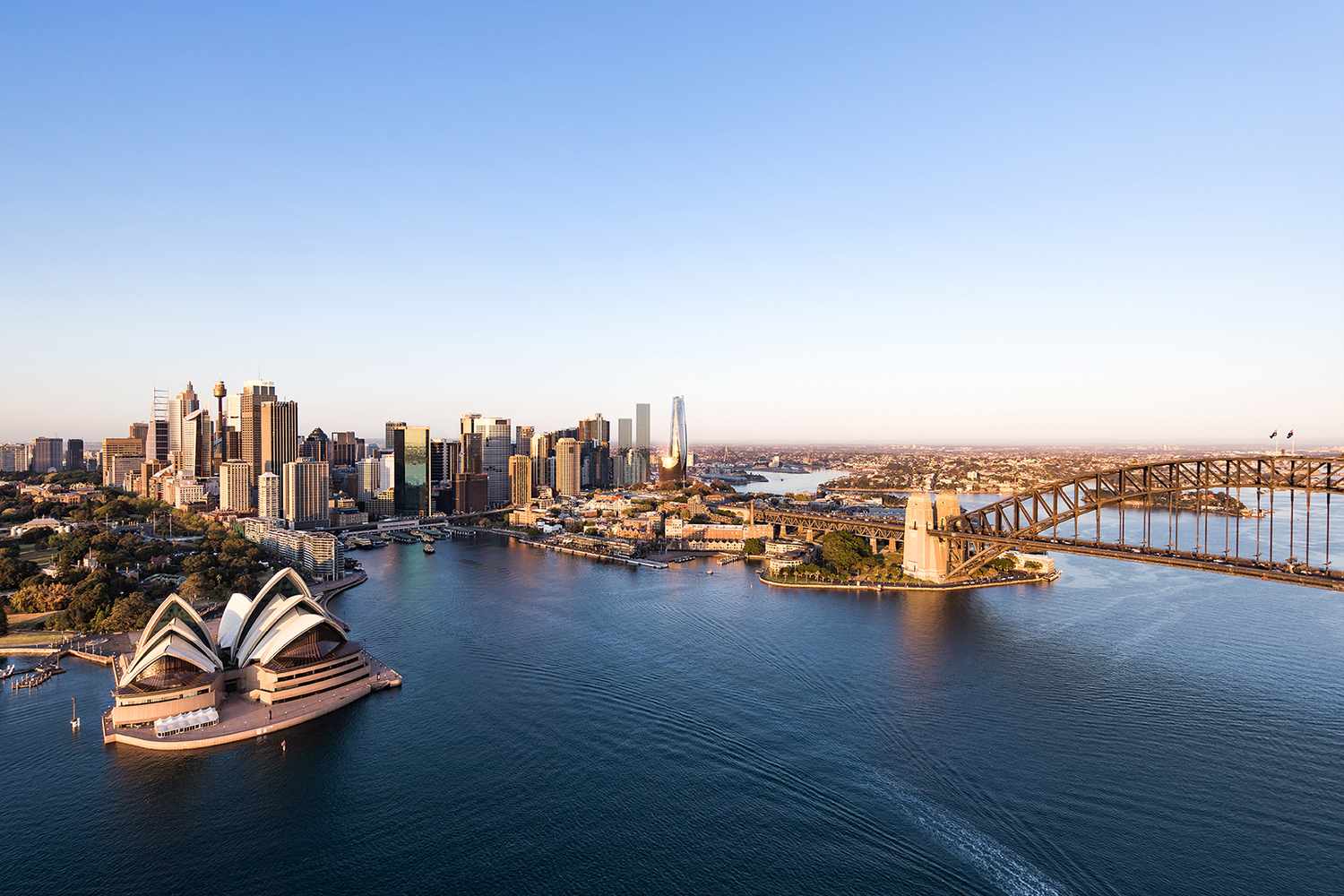}
\caption{View of Sydney (Australia).}
\label{fig:sydney}
\end{figure}

We perform a similar experiment with the image in Figure~\ref{fig:sydney} (correctly located in Sydney, Australia, with 88.1\% confidence) and display the results in Figure~\ref{fig:CAMs}-right. We observe that all of these methods highlight regions relevant to the classification of the images, but only Combi-CAM and Layer-CAM are able to precisely localize these regions. For both images, Combi-CAM highlights significant landmarks of the scenery, such as the Eiffel Tower and other relevant buildings in the Paris image, and the Opera House in the Sydney's image. 
The results of Layer-CAM are also well localized, but our method highlights more regions semantically meaningful.

\begin{figure}[!htb]
\centering
\addtolength{\tabcolsep}{-0.4em}
\begin{tabular}{ccc}
\raisebox{3em}{\footnotesize Grad-CAM} & 
\includegraphics[width=0.35\linewidth, height=0.224\linewidth]{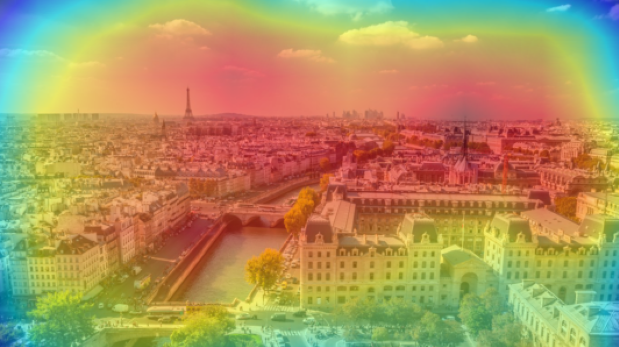} &
\includegraphics[width=0.35\linewidth, height=0.224\linewidth]{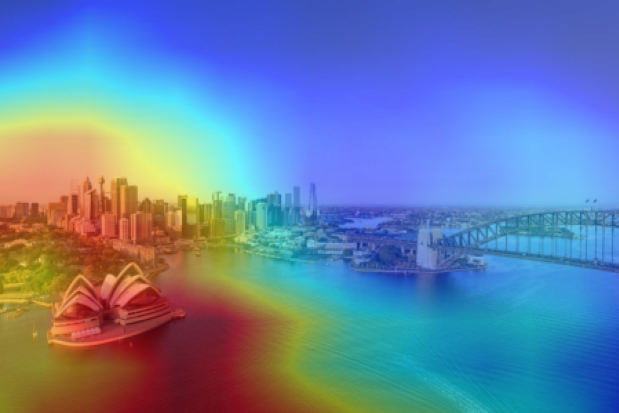} \\
\raisebox{3em}{\footnotesize Grad-CAM++} & 
\includegraphics[width=0.35\linewidth, height=0.224\linewidth]{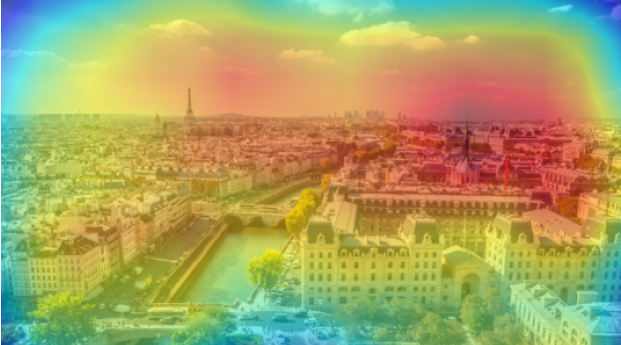} &
\includegraphics[width=0.35\linewidth, height=0.224\linewidth]{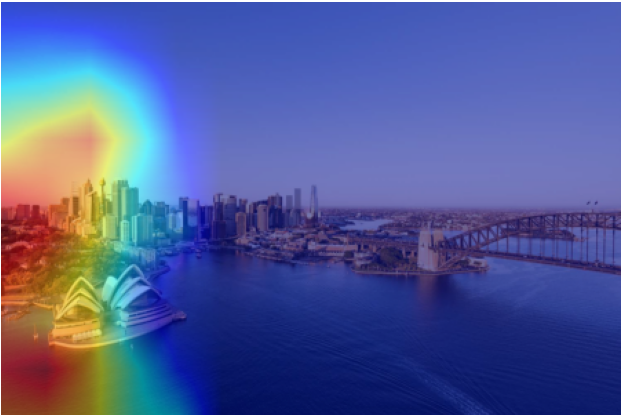} \\
\raisebox{3em}{\footnotesize Layer-CAM} & 
\includegraphics[width=0.35\linewidth, height=0.224\linewidth]{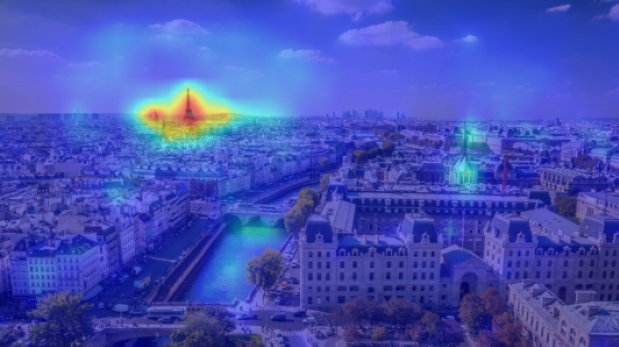} &
\includegraphics[width=0.35\linewidth, height=0.224\linewidth]{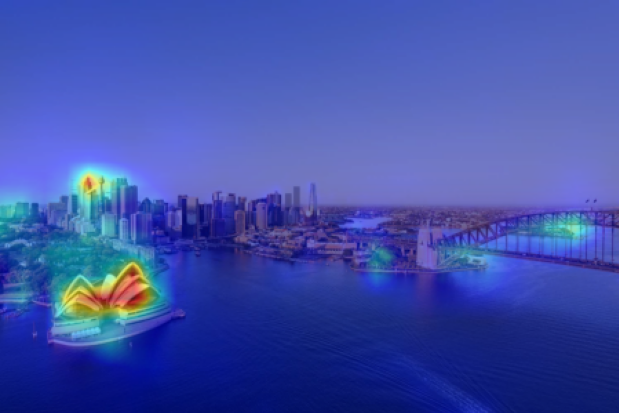} \\
\raisebox{3em}{\footnotesize Score-CAM} & 
\includegraphics[width=0.35\linewidth, height=0.224\linewidth]{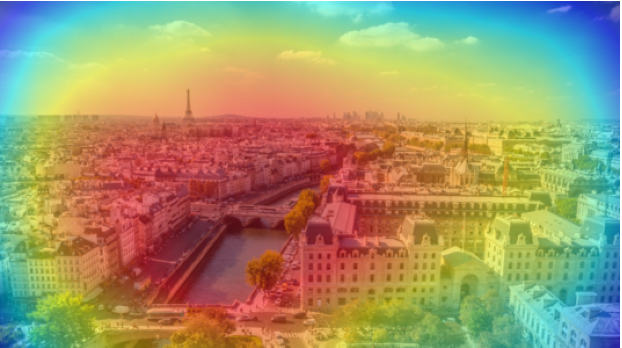} &
\includegraphics[width=0.35\linewidth, height=0.224\linewidth]{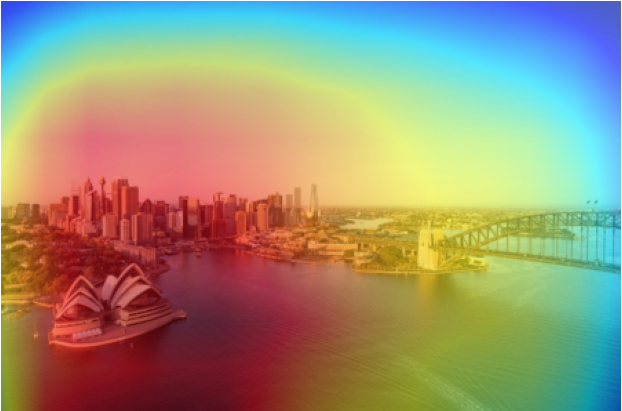} \\
\raisebox{3em}{\footnotesize Combi-CAM} & 
\includegraphics[width=0.35\linewidth, height=0.224\linewidth]{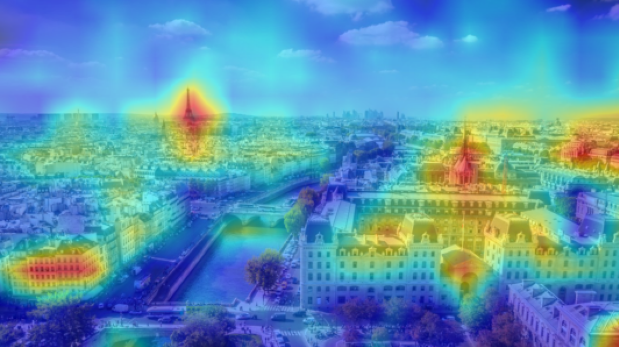} &
\includegraphics[width=0.35\linewidth, height=0.224\linewidth]{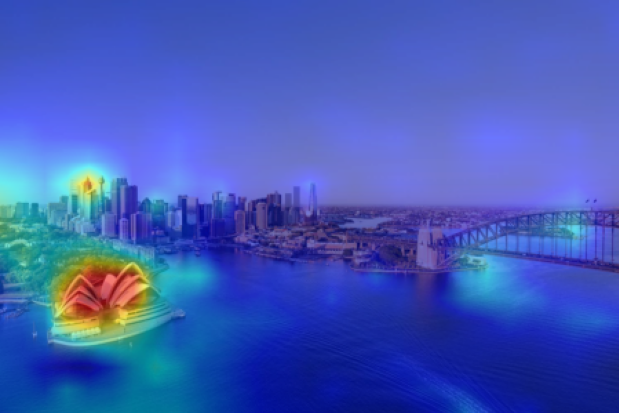}
\end{tabular}
\caption{Activation maps obtained with different methods for the images in Figures~\ref{fig:paris} and~\ref{fig:sydney}. Left column, Paris' image. Right column, Sydney's image.}
\label{fig:CAMs}
\end{figure}

\begin{figure}[!htb]
\centering
\addtolength{\tabcolsep}{-0.4em}
\begin{tabular}{cccc}
\raisebox{3em}{\footnotesize Grad-CAM} & 
\includegraphics[width=0.23\linewidth, height=0.303\linewidth]{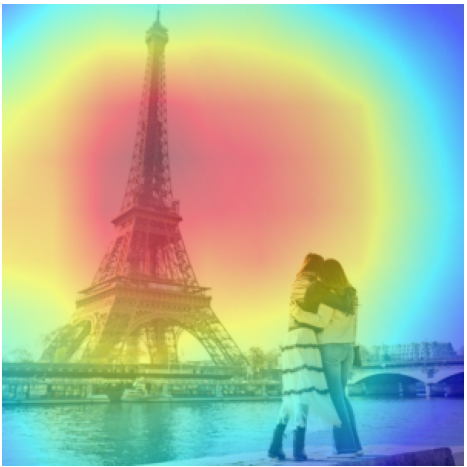} &
\includegraphics[width=0.23\linewidth, height=0.303\linewidth]{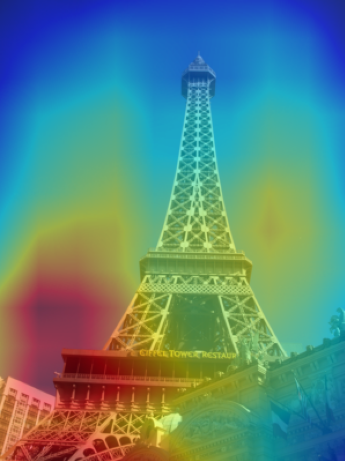} &
\includegraphics[width=0.23\linewidth, height=0.303\linewidth]{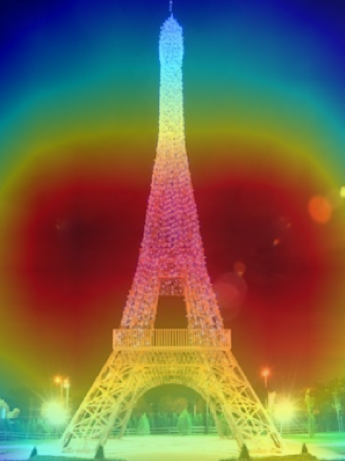} \\
\raisebox{3em}{\footnotesize Grad-CAM++} & 
\includegraphics[width=0.23\linewidth, height=0.303\linewidth]{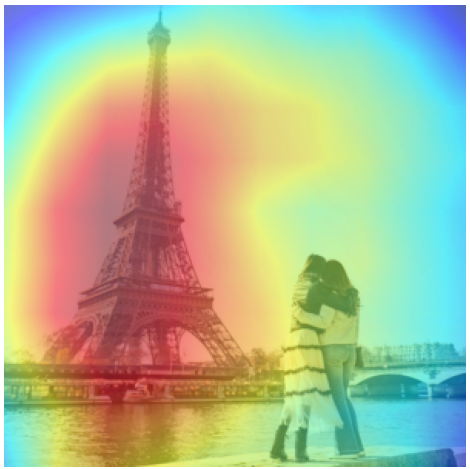} &
\includegraphics[width=0.23\linewidth, height=0.303\linewidth]{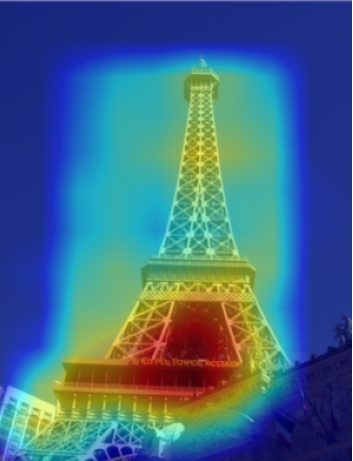} &
\includegraphics[width=0.23\linewidth, height=0.303\linewidth]{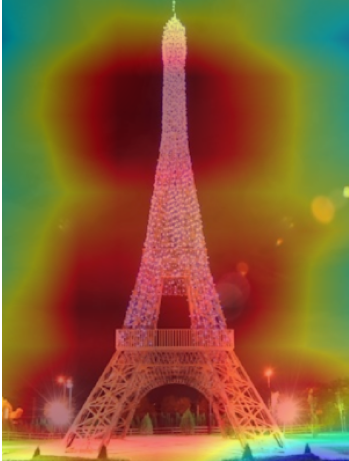} \\
\raisebox{3em}{\footnotesize Layer-CAM} & 
\includegraphics[width=0.23\linewidth, height=0.303\linewidth]{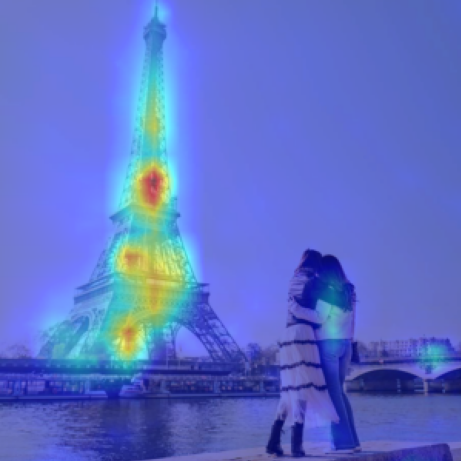} &
\includegraphics[width=0.23\linewidth, height=0.303\linewidth]{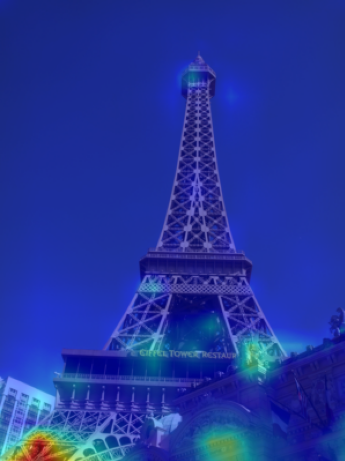} &
\includegraphics[width=0.23\linewidth, height=0.303\linewidth]{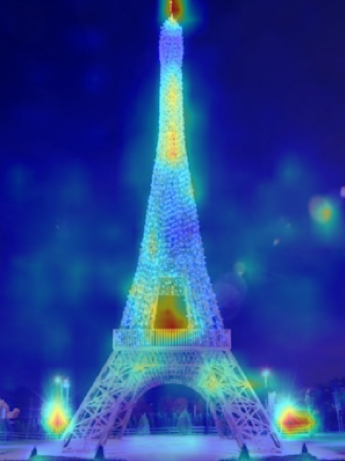} \\
\raisebox{3em}{\footnotesize Score-CAM} & 
\includegraphics[width=0.23\linewidth, height=0.303\linewidth]{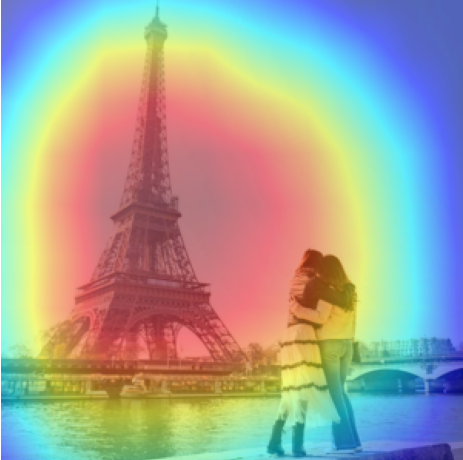} &
\includegraphics[width=0.23\linewidth, height=0.303\linewidth]{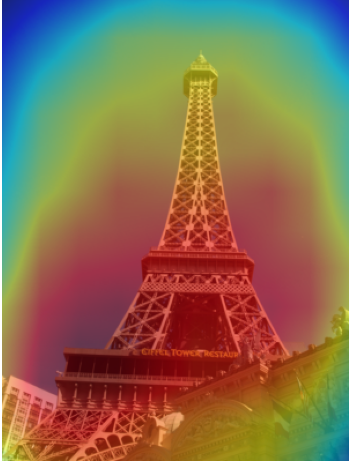} &
\includegraphics[width=0.23\linewidth, height=0.303\linewidth]{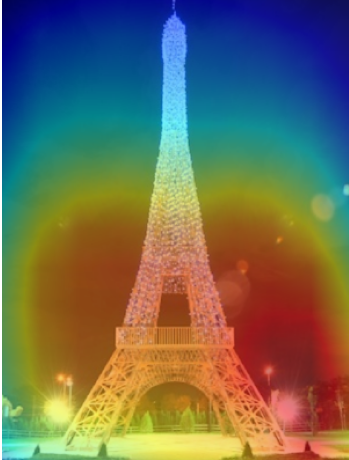} \\
\raisebox{3em}{\footnotesize Combi-CAM} & 
\includegraphics[width=0.23\linewidth, height=0.303\linewidth]{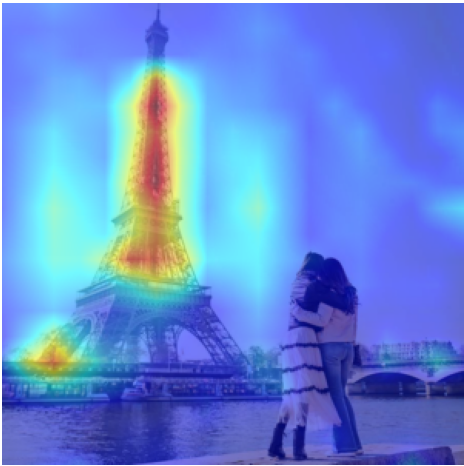} &
\includegraphics[width=0.23\linewidth, height=0.303\linewidth]{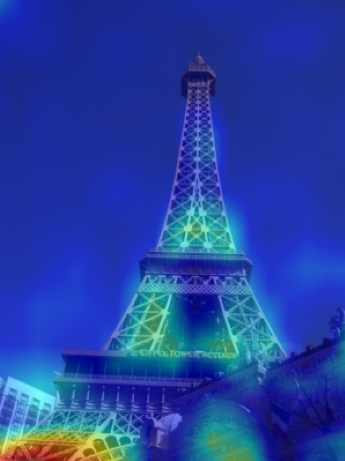} &
\includegraphics[width=0.23\linewidth, height=0.303\linewidth]{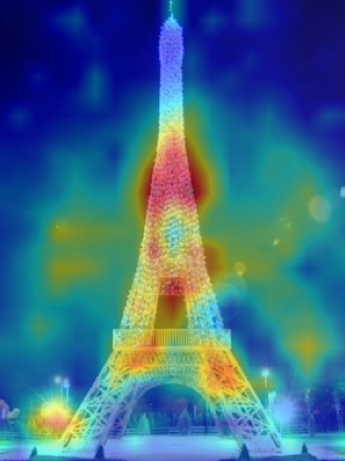}
\end{tabular}
\caption{Activation maps obtained with different methods for the images in Figure~\ref{fig:intro}. Left column, Eiffel tower in Paris; center, replica in Las Vegas; right, replica in Madrid.}
\label{fig:eiffels}
\end{figure}

The example in Figure~\ref{fig:eiffels} shows how Combi-CAM helps to understand why the network fails to geolocalize some images. In this figure, the heatmaps obtained with different attention methods for the images in Figure~\ref{fig:intro} are displayed. The network correctly locates the left and center images in Paris and Las Vegas, but wrongly situates the image on the right also in Paris, although it corresponds to Madrid.
By observing the activation maps we see that Combi-CAM is able to pin-point the regions in the images that lead to the geolocalization results. While for the Las Vegas image contextual clues are taken into account, this does not happen for the Madrid image. In this case, the main clues are based on the tower structure, and thus the image is assigned to the location most likely containing this structure (namely Paris). 
Although the other methods highlight relevant regions of the images, they are unable to precisely locate the most distinctive features, with the exception of Layer-CAM. However, Layer-CAM's results are less comprehensive than ours.

Accurate, layer-aware interpretability should not only visualize decisions but also reveal where discriminative evidence appears in the network and how spatial and semantic cues are balanced along the backbone. By aggregating maps from several layers, Combi-CAM shows when the final block ignores mid-level, localized cues and relies too much on broad context. These findings suggest that both the architecture and the training should fuse intermediate representations with deep semantic features to better combine local evidence with global context.

The readers are invited to access our GitHub repository \text{https://github.com/DavidFaget/Combi-CAM/} where they will find a PyTorch implementation of Combi-CAM. Additional examples of application of the method are also provided in the repository.

\section{\uppercase{Conclusion}}
\label{conclusion}
We propose a new method that permits to visualize which areas of an image have more influence in the output of a network model. The method is based on the aggregation of gradient-weighted activation maps from different layers of the network architecture and has been applied to analyze the output of a geolocalization network. Our results show that the proposed approach permits a precise localization of regions of the image relevant to the classification result of the network and improves over other state-of-the-art techniques. In future work, we shall show how the same approach can be used to analyze other architectures.

\section*{\uppercase{Acknowledgements}}

The first and third authors acknowledge support from the project VERification Assisted by Artificial Intelligence (VERA.AI), funded by the Horizon Europe Framework Programme under grant agreement 101070093. The second author’s contribution to this publication is part of the project PID2021-125711OB-I00, funded by MCIN/AEI/10.13039/501100011033/FEDER, EU.

\bibliographystyle{apalike}
{\small
\bibliography{article}}

\end{document}